\documentclass[11pt]{article}

\usepackage[final]{acl}
\usepackage{times}
\usepackage{latexsym}
\usepackage[T1]{fontenc}
\usepackage[utf8]{inputenc}
\usepackage{microtype}
\usepackage{inconsolata}
\usepackage{graphicx}
\usepackage{amsmath}
\usepackage{amssymb}
\usepackage{algorithm}
\usepackage{algpseudocode}

\title{LAND: Learning an Adaptive Number of Draws for Bayesian Best-of-N Selection}

\author{
  \textbf{Koen Oostermeijer\textsuperscript{*\dag}}
  \and
  \textbf{Teryn Jones\textsuperscript{*}}
  \\
  Aleph Alpha Research
  \\[0.5em]
  \textsuperscript{*}Equal contribution
  \\
  \textsuperscript{\dag}Corresponding author:
  \href{mailto:koen.oostermeijer@aleph-alpha-research.com}{koen.oostermeijer@aleph-alpha-research.com}
}

\begin{document}
\maketitle
\begin{abstract}
Best-of-N decoding improves generation quality by sampling multiple candidates and selecting the best, but it allocates equal compute to sources regardless of how much they benefit from additional sampling: the expected benefit is small for sources with low-variance score distributions and potentially much larger for those with high-variance distributions. To address this shortcoming, we propose LAND (Learning an Adaptive Number of Draws), which dynamically allocates samples according to their posterior-predictive marginal value. LAND learns a discrete empirical mixture of source-conditioned score distributions from calibration data, updates its belief about a source as completion scores are observed, and stops sampling when the expected improvement of another draw falls below a shared compute price. Across machine-translation experiments with multiple models and language pairs, LAND achieves better MetricX scores than fixed best-of-N at matched average compute. Conversely, at matched translation quality, LAND requires approximately 20-40\% fewer candidate generations. Adaptive allocation therefore provides a simple way to retain the benefits of best-of-N while substantially improving its compute efficiency.
\end{abstract}

\section{Introduction}

Model-generated data have become an increasingly important component of the training and fine-tuning of large language models (LLMs)~\citep{long-etal-2024-llms,wang-etal-2023-self-instruct}. A larger teacher model can generate training data for a smaller model, thereby transferring some of its capabilities through knowledge distillation. This approach has a long history in machine translation (MT), where sequence-level knowledge distillation uses translations generated by a teacher model as targets for a student model~\citep{kim-rush-2016-sequence}.

Historically, machine translation systems were trained primarily on naturally occurring parallel data collected from multilingual sources and subsequently filtered and aligned~\citep{tiedemann-2012-parallel}. Synthetic data initially served mainly as a supplement to these corpora, most notably through back-translation of monolingual source texts \citep{sennrich-etal-2016-improving,edunov-etal-2018-understanding}. Over time, however, the balance has shifted: as the quality of machine-translated data outpaced that of web-crawled corpora \citep{finkelstein-etal-2024-introducing}, their share of the training data used by modern MT systems now constitutes an increasingly large fraction, in some cases completely replacing natural data~\citep{finkelstein2026translategemma}. As reliance on generated translations grows, selecting high-quality candidates becomes an increasingly important part of the data-generation pipeline.

One simple way to improve the quality of generated translations is to sample multiple candidates and retain the candidate with the highest score according to some quality metric, a technique referred to as \emph{best-of-$N$} sampling. Previous work has shown that filtering data for quality-estimation metrics can result in improved downstream performance~\citep{peter-etal-2023-theres}.

\begin{figure}[ht]
    \centering
    \includegraphics[width=\columnwidth]{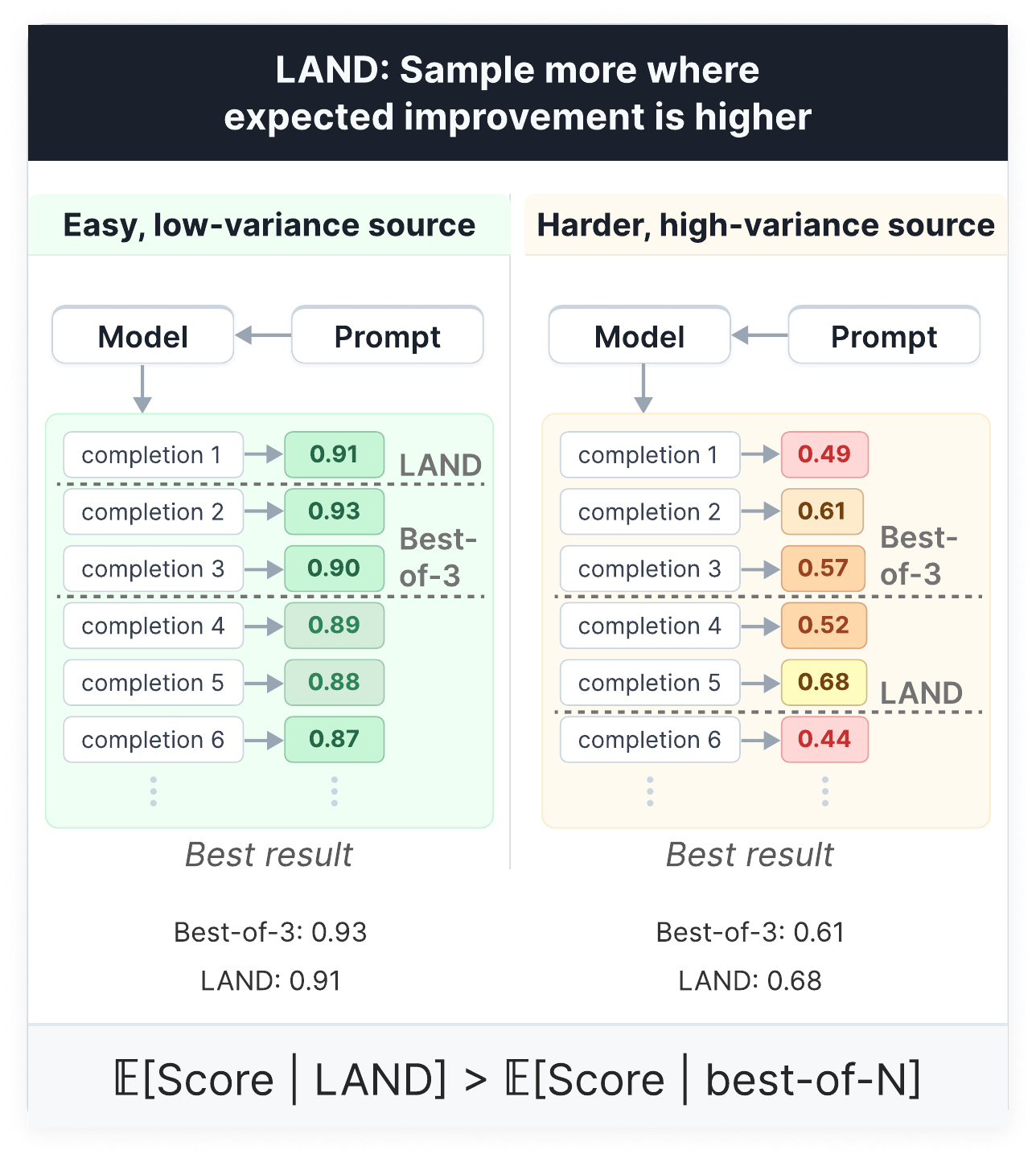}
    \caption{Fixed best-of-N always generates the same number of candidates for every source. However, higher overall scores can be achieved by allocating less compute to sources whose candidate scores are tightly concentrated and more compute to sources with higher-variance score distributions, where additional draws offer greater potential for improvement.}
    \label{fig:land-intuition}
\end{figure}

However, conventional best-of-N generates the same number of candidates for every source sentence. This allocation implicitly assumes that an additional sample is equally valuable for all source sentences. In practice, the distributions of candidate scores can differ substantially between source sentences. As illustrated in Figure~\ref{fig:land-intuition}, some intrinsically simple sentences consistently produce high-scoring translations, such that additional sampling offers little added improvement, while others produce more variable candidates and may benefit considerably from further sampling. Given a collection of source sentences and a fixed generation budget, the relevant question is therefore not only how many candidates to generate, but also where additional candidates are most valuable. This motivates our central research question: how should a fixed generation budget be allocated across source sentences so that additional candidates are generated where they offer the greatest expected marginal improvement?

\begin{figure}[ht]
    \centering
    \includegraphics[width=\columnwidth]{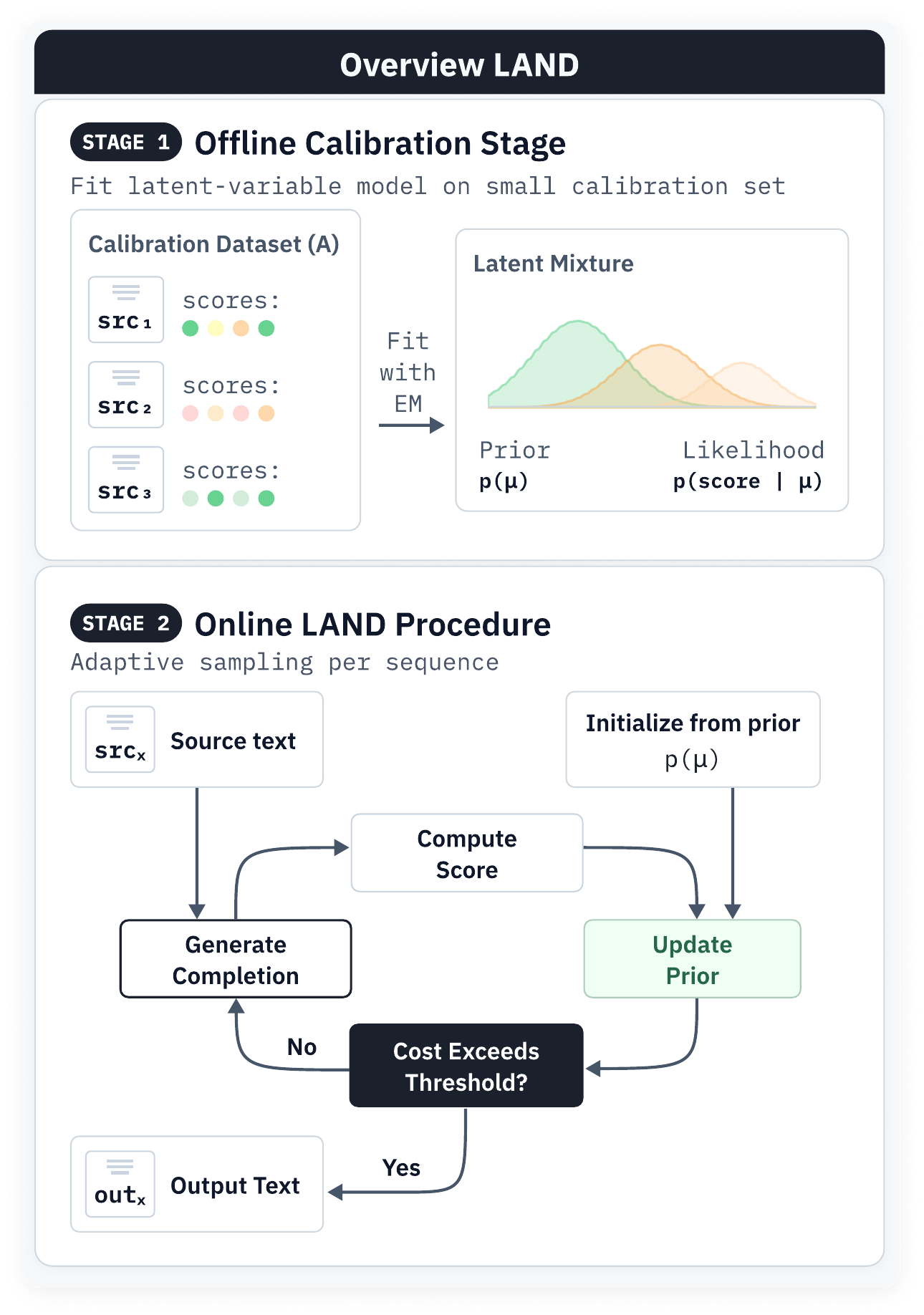}
    \caption{During calibration, LAND fits a discrete empirical mixture and calibrates a global compute price. At deployment, candidates are generated and scored sequentially, the posterior over source types is updated, and sampling continues while the expected improvement of another draw exceeds the compute price.}
    \label{fig:land-overview}
\end{figure}

We introduce \textbf{LAND} (\textbf{L}earning an \textbf{A}daptive \textbf{N}umber of \textbf{D}raws), an adaptive procedure that allocates generation compute according to the expected marginal improvement of additional candidates for each source sentence. A high-level overview of LAND is shown in Figure~\ref{fig:land-overview}. LAND is related in spirit to Weitzman’s Pandora's-box rule \citep{weitzman1978optimal}, which uses reservation values to balance the potential benefit of further search against its cost. Unlike the classical setting, however, the source-specific score distributions in our problem are unknown and each source sentence may be sampled repeatedly. LAND therefore updates a posterior over the score distribution after every observed candidate and adjusts its stopping threshold on the fly. To avoid imposing a restrictive parametric form, we estimate these distributions using a discrete empirical model constructed from observed scores.

In summary, our contributions are as follows:
\begin{itemize}
\setlength{\itemsep}{1pt}
\setlength{\parskip}{0pt}
\setlength{\parsep}{0pt}
\item We formulate best-of-N selection as a compute-allocation problem in which the number of candidates may vary across source sentences under a shared average budget.
\item We derive an oracle allocation rule that relates the decision to generate another candidate to its expected marginal improvement.
\item We introduce LAND, a posterior-adaptive procedure that estimates this improvement online using a discrete empirical model of source-conditioned score distributions.
\item We evaluate LAND against fixed-allocation and adaptive baselines for machine translation under matched average generation budgets.
\end{itemize}

\section{Related Work}

\subsection{Quality Filtering and Selection in Machine Translation}

Quality-estimation metrics have been used both to select translation hypotheses and to filter training data. \citet{fernandes-etal-2022-quality} apply reference-free metrics to N-best reranking and minimum Bayes risk decoding, while \citet{peter-etal-2023-theres} use them to retain higher-quality parallel sentence pairs. For synthetic data, \citet{xu-etal-2022-synthetic} show that the usefulness of back-translated examples depends on both translation quality and example importance. Complementing quality-based filtering, \citet{kocmi-etal-2025-command} prioritize difficult training examples. Related work also uses quality estimates to allocate computation: \citet{cheng-etal-2025-bayesian} selectively evaluate candidates from a fixed list, and \citet{farinhas-etal-2025-translate} defer low-quality translations to a larger model. 

\subsection{Bayesian Stopping and Adaptive Allocation}

LAND shares similarities with Weitzman's Pandora's-box problem, in which costly alternatives with known reward distributions are inspected until further search is no longer worthwhile \citep{weitzman1978optimal}. This framework has subsequently been applied to settings such as cost-aware Bayesian optimization and adaptive LLM sampling \citep{xie2024cost}. Related optimal-stopping approaches have also been proposed for LLM inference. \citet{wan2025beacon} introduce BEACON, which models each prompt’s reward distribution as Gaussian and applies a Bayesian finite-horizon stopping rule, whereas \citet{kalayci2025optimal} learn an unknown Weitzman threshold using upper confidence bounds.

More broadly, LAND is related to Bayesian adaptive-allocation methods such as Thompson sampling~\citep{william1933likelihood}, which use posterior uncertainty to allocate observations sequentially rather than committing the same sampling budget to every alternative.

AdaBoN \citep{raman2025adabon} also studies adaptive Best-of-$N$ allocation, but is designed for batched allocation with relatively large exploration budgets. Since LAND targets sequential stopping in a substantially lower-budget regime, we consider the methods complementary rather than directly comparable.

LAND differs from these approaches in both how it models uncertainty and how it allocates compute. It learns a discrete empirical mixture of source-conditioned score distributions and updates a posterior over source types as candidates are observed. Rather than treating each source independently with a fixed sampling cost, LAND uses a shared compute price to allocate a fixed average generation budget toward sources with the greatest posterior-predictive marginal value of additional draws.


\section{Problem Formulation}
We consider a set of source texts and a fixed compute budget for generating completions. For each source text, an LLM generates one or more candidates, from which the highest-scoring completion is selected.  Metrics that assign lower scores to better candidates are transformed so that higher values consistently correspond to better completions. The budget is specified by a target expected number of completions per source text, $N$, although in practice we allow the realized average to differ slightly due to finite-sample effects.

Our goal is to determine how to allocate the available compute across source texts so as to maximize the average score of the selected completions.

Applying the scoring function to the distribution of generated candidates induces a corresponding distribution over quality scores. Each source text has a latent mean score $\mu$, defined as the expected score of a completion generated for that source text. Across source texts, these means follow the distribution
\begin{align}
\mu \sim \mathcal{D}_{\mathrm d},\label{Pd}
\end{align}
whose expectation value $\mathbb{E}[\mu]$ is the score associated with the best-of-1 baseline, i.e. the average score obtained when exactly one completion is generated per source text.

Conditional on $\mu$, the score $X$ of an individual completion follows a distribution
\begin{align}
X \sim \mathcal{D}_{\mathrm c}\left(\left.\cdot\,  \right| \mu,\sigma^2(\mu)\right),
\end{align}
where $\sigma^2(\mu)$ captures variation among completions generated for the same source text, which may depend on the latent mean $\mu$.


\section{Budget Allocation}

Rather than generating a fixed number of candidates for every source text, i.e. best-of-N, our strategy is to sample until the score exceeds a source-specific threshold. These thresholds are optimized to maximize the average selected score subject to an expected sampling budget of $N$ completions per source text.

In the following we will derive an allocation rule assuming that $\mu$ is known, which is not the case in practice, as this is a latent variable; we relax this assumption in the following section.

For a source text with mean score $\mu$, let $\{X_1,X_2,\ldots\}$ be independent draws from the conditional score distribution $\mathcal{D}_{\mathrm c}(\cdot\mid\mu)$, with probability density function $f_\mu$ and cumulative distribution function $F_\mu$. We define the stopping time $K$ as the number of samples it takes to exceed the threshold $\lambda(\mu)$. Since each draw has a probability of $1-F_\mu(\lambda(\mu))$ to exceed the threshold, $K$ follows a geometric distribution with expectation value
\begin{align}
\mathbb{E}[K\mid\mu]
=
\frac{1}{
1-F_\mu(\lambda(\mu))
}.
\end{align}

Because the sampling process is halted when the stopping condition is met, the final score $X_K$ is also the one corresponding to the selected completion. Its expected value is the threshold plus the expected overshoot:
\begin{align}
    \mathbb{E}[X_K\mid\mu] &= \lambda + \mathbb{E}\left[X-\lambda | X>\lambda,\mu\right]\\
    &= \mathbb{E}\left[X| X>\lambda,\mu\right].
\end{align}

The optimal threshold function maximizes the expected selected score across source texts, subject to an average sampling budget of $N$:
\begin{align}
&\max_{\lambda(\cdot)}\,
\mathbb{E}_{\mu}\left[\mathbb{E}\left[X| X>\lambda(\mu),\mu\right]\right]\nonumber\\
&\text{subject to }
\mathbb{E}_{\mu}
\left[\frac{1}{1-F_\mu(\lambda(\mu))}\right] = N.
\label{eq}
\end{align}

As derived in Appendix~\ref{app}, the optimal threshold satisfies
\begin{align}
    H_\mu(\lambda) \equiv \mathbb{E}\left[\left.\bigl(X-\lambda(\mu)\bigr)_+\right|\mu\right] = \Lambda,
\end{align}
where $(z)_+\equiv\max(z,0)$ denotes the positive part of $z$. Thus, the left-hand side is the expected improvement of one additional draw over the current threshold. The global constant $\Lambda$ is chosen so that the average-budget constraint holds. The optimal policy therefore equalizes the expected marginal improvement across source texts.

\section{Posterior-Adaptive Allocation}
The optimal allocation derived above assumes access to the latent source-specific mean $\mu$. In practice, $\mu$ is unknown and must be inferred from the completion scores observed during sampling. We therefore maintain a posterior distribution over $\mu$ and update the stopping rule after each generated completion.

Let $\mathcal{X}_t \equiv \{X_1,\ldots,X_t\}$ denote the scores observed after $t$ draws for a given source text, and let
\begin{align}
q_t(\mu)
\equiv
p(\mu\mid\mathcal{X}_t)
\end{align}
be the corresponding posterior over its latent mean. Before observing any completions, this posterior is initialized as the population-level prior from Equation~\ref{Pd},
\begin{align}
q_0(\mu) = P_{\mathrm d}(\mu).
\end{align}
Then, after observing a new score $X_{t+1}$, Bayes' rule gives
\begin{align}
q_{t+1}(\mu) = \frac{ f_\mu(X_{t+1})q_t(\mu)}{\int\mathrm{d}\mu'\, f_{\mu'}(X_{t+1})q_t(\mu')}
\label{eq}
\end{align}
where $f_\mu$ is the conditional score density for a source text with mean $\mu$.

The expected-improvement function is averaged over this posterior:
\begin{align}
    \widehat H_t(\lambda) &\equiv \int\mathrm{d}\mu\,q_t(\mu)H_\mu(\lambda)\\
    &= \mathbb{E}\left[\left.(X_{t+1}-\lambda)_+\right|\mathcal{X}_t\right].
\end{align}
Thus, $\widehat H_t(\lambda)$ is the expected improvement of one additional draw over a current score $\lambda$.

Let $B_t \equiv \max_{1\leq i\leq t}X_i$ be the best score observed so far. The optimal policy is to continue sampling whenever it exceeds the global compute cost $\Lambda$:
\begin{align}
\widehat H_t(B_t)>\Lambda.
\end{align}
Equivalently, we may define an adaptive threshold $\lambda_t$ by
\begin{align}
\widehat H_t(\lambda_t)=\Lambda.
\end{align}
The procedure continues while $B_t<\lambda_t$. Unlike the previous threshold $\lambda(\mu)$, the adaptive threshold $\lambda_t$ changes as new scores are observed. Each completion updates both the best score $B_t$ and the posterior belief about the source-specific score distribution, allowing the stopping decision to adapt online.

\section{Practical Implementation}
\label{sec:land-practical}

The practical implementation of LAND starts by estimating the score distributions using a small calibration set. We first rescale all scores to the interval $[0,1]$, with larger values denoting better completions. These scores are then partitioned into $J$ cells, with each cell $j$ assigned a representative value $v_j$. The boundaries of these cells are initialized using the quantiles of these calibration scores, which provides greater resolution in densely populated regions of the score distribution.

We model variation across source texts using a mixture of $I$ latent source distributions. We do not assume that source variation is fundamentally discrete. Rather, this is a finite approximation to continuous population heterogeneity while reducing posterior updates and expected-improvement calculations to finite sums. Each type $i$ has prior probability
\begin{align}
\pi_i = \Pr(z=i),
\end{align}
and a categorical distribution over score cells,
\begin{align}
p_{ij}
=
\Pr(X \text{ falls in cell } j | z=i).
\end{align}
Here, $z$ denotes the latent source type and $X$ denotes the score of a generated completion.  This discretized distribution greatly simplifies calculations and avoids imposing a restrictive parametric form on the source-conditioned score distributions. In particular, we found that parametric families such as (truncated) Gaussian or Beta distributions often impose inaccurate assumptions about tail behavior, which can substantially distort LAND's cost allocation policy.

Since the latent source types are not observed, we fit the mixture using expectation-maximization (EM). Suppose the calibration set contains $L$ source texts. For source $\ell$, let $n_{\ell j}$ denote the number of observed completion scores that fall in cell $j$, and let
\begin{align}
n_\ell = \sum_{j=1}^{J} n_{\ell j}.
\end{align}
EM estimates the model parameters by maximizing the marginal log-likelihood of the observed data:
\begin{align}
\mathcal{L}
=
\sum_{\ell=1}^{L}
\log
\left[
\sum_{i=1}^{I}
\pi_i
\prod_{j=1}^{J}
p_{ij}^{n_{\ell j}}
\right].
\end{align}

After fitting the mixture, we use it to initialize the stopping thresholds: For a source of latent type $i$, the expected improvement from one additional draw over a current best score $b$ is computed as
\begin{align}
H_i(b)
=
\sum_{j=1}^{J}
p_{ij}(v_j-b)_+.
\end{align}

During sampling, each component-specific expected improvement is weighted by the posterior probability that the source belongs to the corresponding latent type. These weights are initialized as the population mixture weights, $q_{0,i}=\pi_i$. After observing a completion score $x_{t+1}$ in cell $c(x_{t+1})$, the posterior is updated as
\begin{align}
q_{t+1,i}
=
\frac{
q_{t,i}
p_{i,c(x_{t+1})}^{\tau}
}{
\sum_{k=1}^{I}
q_{t,k}
p_{k,c(x_{t+1})}^{\tau}
},
\end{align}
where $\tau\in(0,1]$ is a likelihood-tempering parameter. The standard Bayesian update is recovered when $\tau=1$. We use tempering to prevent the posterior from becoming overly concentrated after only a small number of observations.

Let $B_t$ denote the best score observed after $t$ draws. The posterior-predictive expected improvement of one additional draw is
\begin{align}
\widehat H_t(B_t)
=
\sum_{i=1}^{I}
q_{t,i}H_i(B_t).
\end{align}
LAND continues sampling whenever this expected improvement exceeds the global compute price $\Lambda$. In addition, we impose a maximum number of draws per source text of $K_{\max}$ to bound the worst-case allocation.

\subsection{Calibrating $\Lambda$}
The global cost limit $\Lambda$ is calibrated by replaying the complete posterior-adaptive policy on a held-out set of calibration trajectories. For a candidate value of $\Lambda$, each trajectory is processed using the same posterior updates, running maximum draw rule, and sampling cap used at deployment. If $K_\ell(\Lambda)$ denotes the resulting number of draws for replay source $\ell$, the estimated average compute is set to equal $N$:
\begin{align}
\frac{1}{L_{\mathrm{replay}}}
\sum_{\ell=1}^{L_{\mathrm{replay}}}
K_\ell(\Lambda) \approx N.
\end{align}
This is to ensure the policy captures posterior transitions and stopping behavior as closely as possible.

Finally, to correct for any residual calibration error or potential distribution shift, we update $\Lambda$ after each batch of completed source texts. Let $\overline K_m$ be the average number of draws in batch $m$. We apply the log-space update
\begin{align}
\log\Lambda_{m+1} = \log\Lambda_m + \eta \frac{\overline K_m-N}{N}.
\end{align}
When the realized number of completions exceeds the target budget, the update increases $\Lambda$ and makes further sampling less likely, and vice versa when the realized number falls below the target.

\subsection{Complete LAND Procedure}
The preceding subsections describe the individual components of LAND: fitting the discrete empirical mixture, updating the posterior over latent source types, estimating the posterior-predictive value of another draw, and calibrating the global compute price. The procedure is sequential: once LAND stops sampling a source and moves to the next source, that source is not revisited. Algorithm~\ref{alg} summarizes how these components are combined into the complete procedure. The mixture model and initial value of $\Lambda$ are estimated during calibration. At deployment, each newly observed completion updates both the current best score and the posterior over source types, and sampling continues only while the posterior-predictive expected improvement justifies its compute cost, subject to the maximum-draw constraint.

\begin{algorithm}[t]
\caption{LAND with a discrete empirical mixture}
\label{alg}
\begin{algorithmic}[1]

\Require Mixture-fit trajectories $\mathcal C_{\mathrm{fit}}$,
replay trajectories $\mathcal C_{\mathrm{replay}}$,
target budget $N$, tempering $\tau$, cap $K_{\max}$

\State Construct score cells and representatives $\{v_j\}_{j=1}^{J}$
\State Fit $\{\pi_i,p_{ij}\}$ on $\mathcal C_{\mathrm{fit}}$ using EM
\State Compute $H_i(b)=\sum_j p_{ij}(v_j-b)_+$
\State Calibrate $\Lambda$ on $\mathcal C_{\mathrm{replay}}$

\For{each deployment source $s$}
    \State $q_i \gets \pi_i$, $B \gets -\infty$, $K \gets 0$
    \Repeat
        \State Generate and score one completion $x$
        \State $K \gets K+1$, $B \gets \max(B,x)$
        \State $q_i \gets
        q_i p_{i,c(x)}^\tau /
        \sum_k q_k p_{k,c(x)}^\tau$
    \Until{$K \geq K_{\max}$ or $\sum_i q_i H_i(B) \leq \Lambda$}
    \State Return the highest-scoring observed completion
    \State Add $K$ to the current update batch
    \If{the update batch is full}
        \State Update $\Lambda$ using the batched log-space rule
    \EndIf
\EndFor

\end{algorithmic}
\end{algorithm}

\section{Results}

We evaluate LAND on source sentences drawn from the OPUS parallel corpus.~\citep{tiedemann-2012-parallel} We use only the source side of each selected sentence pair and generate new translations with the evaluated model; the original target sentences are not used as generation targets. Using Gemma4 31B,~\citep{team2026gemma} we consider four translation directions: Ukrainian$\rightarrow$German, Spanish$\rightarrow$English, Dutch$\rightarrow$Hungarian, and Turkish$\rightarrow$Russian. Translation quality is evaluated with reference-free MetricX\footnote{\url{https://huggingface.co/google/metricx-24-hybrid-xxl-v2p6-bfloat16}}~\citep{juraska-etal-2024-metricx}. For each translation direction and target budget $N$, we separately tune the sampling cap $K_{\max}$ and tempering parameter $\tau$ using the corresponding mixture-fit and replay trajectories. We evaluate average sampling budgets up to $N=32$ and compare LAND against fixed best-of-$N$ selection at each budget. Using MetricX for both selection and evaluation is intentional: LAND addresses how efficiently a given filtering objective can be optimized, rather than whether that objective improves downstream model training. The utility of QE-based filtering for MT data has been established independently by prior work \citep{peter-etal-2023-theres}.

\begin{figure}[t]
    \centering
    \includegraphics[width=\columnwidth]{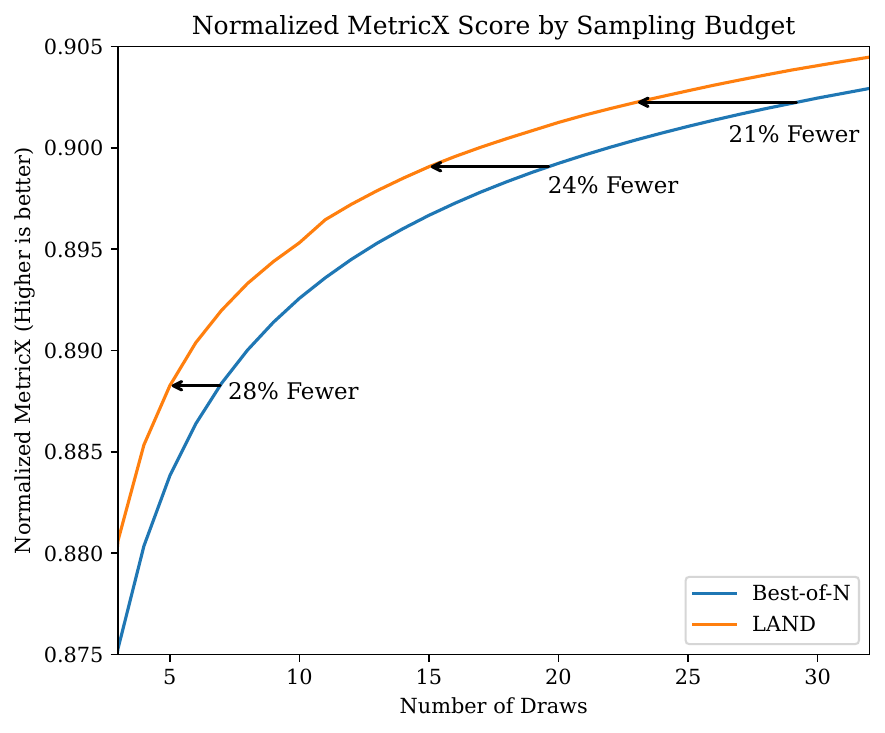}
    \caption{
    Normalized MetricX score as a function of the number of draws, averaged over the four Gemma4 31B translation directions. LAND reaches the same translation quality as fixed best-of-N using approximately 20-30\% fewer draws at the highlighted operating points.
    }
    \label{fig:scores}
\end{figure}

Figure~\ref{fig:scores} shows the normalized MetricX score as a function of the sampling budget, averaged across the four Gemma4 language pairs. LAND consistently achieves higher translation quality than fixed best-of-N at the same number of draws. Equivalently, LAND reaches the same MetricX score with substantially fewer candidate generations. At the operating points highlighted in the figure, LAND matches the performance of fixed best-of-N while using approximately 20-30\% fewer draws. Although the relative savings are largest at lower budgets, they remain substantial throughout the evaluated range.

\begin{figure}[t]
    \centering
    \includegraphics[width=\columnwidth]{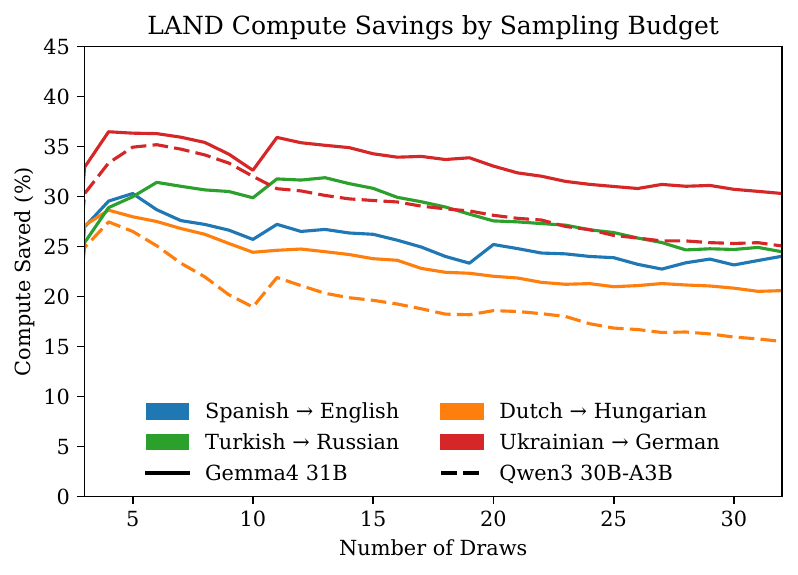}
    \caption{
    Candidate-generation savings of LAND relative to fixed best-of-N at matched MetricX performance.  LAND reduces the number of required draws across all evaluated language pairs and sampling budgets.
    }
    \label{fig:savings}
\end{figure}

Figure~\ref{fig:savings} reports the candidate-generation savings separately for each translation direction. In addition to the four Gemma4 31B experiments, we evaluate Ukrainian$\rightarrow$German and Dutch$\rightarrow$Hungarian using Qwen3 30B-A3B~\citep{qwen3technicalreport}. The savings are computed relative to fixed best-of-N at matched MetricX performance. Across models, language pairs, and sampling budgets, LAND reduces the required number of completions by approximately 20-40\%.

The magnitude of the savings varies across translation directions. Ukrainian$\rightarrow$German exhibits the largest and most consistent reductions in compute, while Dutch$\rightarrow$Hungarian generally shows smaller gains, especially for Qwen3 30B-A3B at larger budgets. Nevertheless, LAND provides positive candidate-generation savings across all evaluated settings.

LAND targets an average budget of $N$ rather than enforcing this exactly. The adaptive compute-price update nevertheless keeps the realized budget very close to its target: across the six evaluated model-language settings, the mean absolute deviation from $N$ ranges from 0.01 to 0.04 draws. Across repeated runs on the same data with different random orderings, the standard deviation of the resulting draw savings is approximately 0.8-1.5 percentage points across the evaluated settings.

Overall, these results show that allocating the same number of candidates to every source sentence is inefficient. By directing additional samples toward source sentences for which they have greater posterior-predictive value, LAND preserves the quality gains of best-of-N while substantially reducing generation cost.

\subsection{Comparison to BEACON}
Finally, we compare LAND against BEACON~\citep{wan2025beacon}, a Bayesian adaptive-sampling method that models the reward distribution for each source as Gaussian and updates its parameters online. In our experiments, BEACON underperforms fixed best-of-N, requiring more compute to reach the same average MetricX score for all sampling budgets, language pairs, and generation models considered.  This seems to indicate that its stopping rule is not well suited to the score distributions encountered in this setting. One possible explanation is the Gaussian modeling assumption: the source-conditioned MetricX distributions can exhibit shapes that are poorly captured by a single parametric distribution. LAND's performance similarly suffered under this modeling assumption. A second possible factor is that BEACON must infer the reward distribution of a new source from the small number of samples observed for that source, while LAND can exploit population-level structure learned across calibration sources. However, we have not isolated the contribution of these differences, and a more detailed investigation would be required to determine the cause of BEACON's underperformance in this setting.

\section{Conclusion and Limitations}

We introduced LAND, an adaptive alternative to fixed best-of-N sampling that allocates generation compute according to the posterior-predictive value of additional candidates. Rather than committing to the same number of draws for every source, LAND uses observed completion scores to update its belief about the source-specific score distribution and continues sampling only when the expected improvement of another draw exceeds a global compute price. A discrete empirical mixture provides a flexible approximation to the population of source-conditioned score distributions, while replay-based calibration allows the resulting policy to satisfy a prescribed average generation budget.

Across the machine-translation settings considered in this work, LAND consistently improves the compute-quality trade-off relative to fixed best-of-$N$. At matched average compute, adaptive allocation yields better translation quality; conversely, at matched MetricX performance, LAND reduces the required number of candidate generations by approximately 20-40\% across the evaluated models, language pairs, and sampling budgets. These results indicate that a substantial fraction of best-of-N compute can be saved simply by reallocating samples toward more promising sources.

Although our experiments focus on machine translation, the formulation of LAND is not specific to translation. The method applies whenever multiple candidate outputs can be sampled for the same input, evaluated with a continuous scalar reward, and compared to select the best candidate. It therefore extends directly to other best-of-N settings, although performance will depend on the informativeness of the chosen score. Our current experiments validate LAND only with MetricX, on a limited set of model families, and in machine translation. Future work will test alternative metrics, additional model families and scales, and other generation tasks equipped with continuous reward or quality scores.

Some limitations remain. First, LAND requires calibration data containing multiple completion scores per source in order to estimate the empirical mixture and calibrate the compute price. This introduces an upfront cost that fixed best-of-N does not require and may be undesirable when only a small dataset is to be generated.  Second, the fitted score-distribution model remains fixed during deployment. Although LAND updates its posterior for each individual source and adjusts the compute price $\Lambda$ to control the realized average budget, it does not use the completion scores collected across previously processed deployment sources to refine the empirical mixture itself. Consequently, additional information acquired during deployment is unused. Extending LAND to periodically update its mixture parameters from newly observed scores could reduce its dependence on an initial calibration set and allow it to adapt to distribution shift over time. And finally, we use the number of candidate generations as a proxy for compute. Since generations can differ in length and execution cost, the reported reduction in draws need not translate one-to-one into wall-clock or hardware savings. We leave this to future work.


\bibliography{custom}

@article{finkelstein2026translategemma,
  title={Translategemma technical report},
  author={Finkelstein, Mara and Caswell, Isaac and Domhan, Tobias and Peter, Jan-Thorsten and Juraska, Juraj and Riley, Parker and Deutsch, Daniel and Kovacs, Geza and Dilanni, Cole and Cherry, Colin and others},
  journal={arXiv preprint arXiv:2601.09012},
  year={2026}
}

@book{weitzman1978optimal,
  title={Optimal search for the best alternative},
  author={Weitzman, Martin},
  volume={78},
  number={8},
  year={1978},
  publisher={Department of Energy}
}

@article{wan2025beacon,
  title={Beacon: Bayesian optimal stopping for efficient llm sampling},
  author={Wan, Guangya and Xu, Zixin Stephen and Zorc, Sasa and Baucells, Manel and Hu, Mengxuan and Wang, Hao and Li, Sheng},
  journal={arXiv preprint arXiv:2510.15945},
  year={2025}
}

@article{kalayci2025optimal,
  title={Optimal Stopping vs Best-of-$ N $ for Inference Time Optimization},
  author={Kalayci, Yusuf and Raman, Vinod and Dughmi, Shaddin},
  journal={arXiv preprint arXiv:2510.01394},
  year={2025}
}

@article{xie2024cost,
  title={Cost-aware Bayesian optimization via the pandora's box gittins index},
  author={Xie, Qian and Astudillo, Raul and Frazier, Peter I and Scully, Ziv and Terenin, Alexander},
  journal={Advances in Neural Information Processing Systems},
  volume={37},
  pages={115523--115562},
  year={2024}
}

@article{team2026gemma,
  title={Gemma 4 technical report},
  author={{Gemma Team}},
  journal={arXiv preprint arXiv:2607.02770},
  year={2026}
}

@misc{qwen3technicalreport,
      title={Qwen3 Technical Report}, 
      author={{Qwen Team}},
      year={2025},
      eprint={2505.09388},
      archivePrefix={arXiv},
      primaryClass={cs.CL},
      url={https://arxiv.org/abs/2505.09388}, 
}

@article{william1933likelihood,
  title={On the likelihood that one unknown probability exceeds another in view of the evidence of two samples},
  author={William, R Thompson},
  journal={Biometrika},
  volume={25},
  number={3/4},
  pages={285--294},
  year={1933}
}

@article{raman2025adabon,
  title={AdaBoN: Adaptive Best-of-N Alignment},
  author={Raman, Vinod and Asi, Hilal and Kale, Satyen},
  journal={arXiv preprint arXiv:2505.12050},
  year={2025}
}

@inproceedings{kocmi-etal-2025-command,title = "Command-A-Translate: Raising the Bar of Machine Translation with Difficulty Filtering",author = {Kocmi, Tom and Arkhangorodsky, Arkady and Berard, Alexandre and Blunsom, Phil and Cahyawijaya, Samuel and Dehaze, Th{\'e}o and Fadaee, Marzieh and Frosst, Nicholas and Galle, Matthias and Gomez, Aidan and Govindarajan, Nithya and Ko, Wei-Yin and Kreutzer, Julia and Marchisio, Kelly and {\"U}st{\"u}n, Ahmet and Vincent, Sebastian and Zhang, Ivan},editor = "Haddow, Barry and Kocmi, Tom and Koehn, Philipp and Monz, Christof",booktitle = "Proceedings of the Tenth Conference on Machine Translation",month = nov,year = "2025",address = "Suzhou, China",publisher = acl,url = anth # {2025.wmt-1.55/},doi = "10.18653/v1/2025.wmt-1.55",pages = "789--799",ISBN = "979-8-89176-341-8"}

@inproceedings{cheng-etal-2025-bayesian,title = "A {B}ayesian Optimization Approach to Machine Translation Reranking",author = {Cheng, Julius and Z{\"u}fle, Maike and Zouhar, Vil{\'e}m and Vlachos, Andreas},editor = "Chiruzzo, Luis and Ritter, Alan and Wang, Lu",booktitle = "Proceedings of the 2025 Conference of the Nations of the Americas Chapter of the Association for Computational Linguistics: Human Language Technologies (Volume 1: Long Papers)",month = apr,year = "2025",address = "Albuquerque, New Mexico",publisher = acl,url = anth # {2025.naacl-long.145/},doi = "10.18653/v1/2025.naacl-long.145",pages = "2849--2862",ISBN = "979-8-89176-189-6"}

@inproceedings{farinhas-etal-2025-translate,title = "Translate Smart, not Hard: Cascaded Translation Systems with Quality-Aware Deferral",author = "Farinhas, Ant{\'o}nio and Guerreiro, Nuno M and Agrawal, Sweta and Rei, Ricardo and Martins, Andre",editor = "Christodoulopoulos, Christos and Chakraborty, Tanmoy and Rose, Carolyn and Peng, Violet",booktitle = "Proceedings of the 2025 Conference on Empirical Methods in Natural Language Processing",month = nov,year = "2025",address = "Suzhou, China",publisher = acl,url = anth # {2025.emnlp-main.1358/},doi = "10.18653/v1/2025.emnlp-main.1358",pages = "26742--26756",ISBN = "979-8-89176-332-6"}

@inproceedings{juraska-etal-2024-metricx,title = "{M}etric{X}-24: The {G}oogle Submission to the {WMT} 2024 Metrics Shared Task",author = "Juraska, Juraj and Deutsch, Daniel and Finkelstein, Mara and Freitag, Markus",editor = "Haddow, Barry and Kocmi, Tom and Koehn, Philipp and Monz, Christof",booktitle = "Proceedings of the Ninth Conference on Machine Translation",month = nov,year = "2024",address = "Miami, Florida, USA",publisher = acl,url = anth # {2024.wmt-1.35/},doi = "10.18653/v1/2024.wmt-1.35",pages = "492--504"}

@inproceedings{finkelstein-etal-2024-introducing,title = "Introducing the {N}ews{P}a{LM} {MBR} and {QE} Dataset: {LLM}-Generated High-Quality Parallel Data Outperforms Traditional Web-Crawled Data",author = "Finkelstein, Mara and Vilar, David and Freitag, Markus",editor = "Haddow, Barry and Kocmi, Tom and Koehn, Philipp and Monz, Christof",booktitle = "Proceedings of the Ninth Conference on Machine Translation",month = nov,year = "2024",address = "Miami, Florida, USA",publisher = acl,url = anth # {2024.wmt-1.126/},doi = "10.18653/v1/2024.wmt-1.126",pages = "1355--1372"}

@inproceedings{long-etal-2024-llms,title = "On {LLM}s-Driven Synthetic Data Generation, Curation, and Evaluation: A Survey",author = "Long, Lin and Wang, Rui and Xiao, Ruixuan and Zhao, Junbo and Ding, Xiao and Chen, Gang and Wang, Haobo",editor = "Ku, Lun-Wei and Martins, Andre and Srikumar, Vivek",booktitle = "Findings of the Association for Computational Linguistics: ACL 2024",month = aug,year = "2024",address = "Bangkok, Thailand",publisher = acl,url = anth # {2024.findings-acl.658/},doi = "10.18653/v1/2024.findings-acl.658",pages = "11065--11082"}

@inproceedings{peter-etal-2023-theres,title = "There{'}s No Data like Better Data: Using {QE} Metrics for {MT} Data Filtering",author = "Peter, Jan-Thorsten and Vilar, David and Deutsch, Daniel and Finkelstein, Mara and Juraska, Juraj and Freitag, Markus",editor = "Koehn, Philipp and Haddow, Barry and Kocmi, Tom and Monz, Christof",booktitle = "Proceedings of the Eighth Conference on Machine Translation",month = dec,year = "2023",address = "Singapore",publisher = acl,url = anth # {2023.wmt-1.50/},doi = "10.18653/v1/2023.wmt-1.50",pages = "561--577"}

@inproceedings{wang-etal-2023-self-instruct,title = "Self-Instruct: Aligning Language Models with Self-Generated Instructions",author = "Wang, Yizhong and Kordi, Yeganeh and Mishra, Swaroop and Liu, Alisa and Smith, Noah A. and Khashabi, Daniel and Hajishirzi, Hannaneh",editor = "Rogers, Anna and Boyd-Graber, Jordan and Okazaki, Naoaki",booktitle = "Proceedings of the 61st Annual Meeting of the Association for Computational Linguistics (Volume 1: Long Papers)",month = jul,year = "2023",address = "Toronto, Canada",publisher = acl,url = anth # {2023.acl-long.754/},doi = "10.18653/v1/2023.acl-long.754",pages = "13484--13508"}

@inproceedings{xu-etal-2022-synthetic,title = "On Synthetic Data for Back Translation",author = "Xu, Jiahao and Ruan, Yubin and Bi, Wei and Huang, Guoping and Shi, Shuming and Chen, Lihui and Liu, Lemao",editor = "Carpuat, Marine and de Marneffe, Marie-Catherine and Meza Ruiz, Ivan Vladimir",booktitle = "Proceedings of the 2022 Conference of the North American Chapter of the Association for Computational Linguistics: Human Language Technologies",month = jul,year = "2022",address = "Seattle, United States",publisher = acl,url = anth # {2022.naacl-main.32/},doi = "10.18653/v1/2022.naacl-main.32",pages = "419--430"}

@inproceedings{fernandes-etal-2022-quality,title = "Quality-Aware Decoding for Neural Machine Translation",author = "Fernandes, Patrick and Farinhas, Ant{\'o}nio and Rei, Ricardo and C. de Souza, Jos{\'e} G. and Ogayo, Perez and Neubig, Graham and Martins, Andre",editor = "Carpuat, Marine and de Marneffe, Marie-Catherine and Meza Ruiz, Ivan Vladimir",booktitle = "Proceedings of the 2022 Conference of the North American Chapter of the Association for Computational Linguistics: Human Language Technologies",month = jul,year = "2022",address = "Seattle, United States",publisher = acl,url = anth # {2022.naacl-main.100/},doi = "10.18653/v1/2022.naacl-main.100",pages = "1396--1412"}

@inproceedings{edunov-etal-2018-understanding,title = "Understanding Back-Translation at Scale",author = "Edunov, Sergey and Ott, Myle and Auli, Michael and Grangier, David",editor = "Riloff, Ellen and Chiang, David and Hockenmaier, Julia and Tsujii, Jun{'}ichi",booktitle = "Proceedings of the 2018 Conference on Empirical Methods in Natural Language Processing",month = oct # "-" # nov,year = "2018",address = "Brussels, Belgium",publisher = acl,url = anth # {D18-1045/},doi = "10.18653/v1/D18-1045",pages = "489--500"}

@inproceedings{sennrich-etal-2016-improving,title = "Improving Neural Machine Translation Models with Monolingual Data",author = "Sennrich, Rico and Haddow, Barry and Birch, Alexandra",editor = "Erk, Katrin and Smith, Noah A.",booktitle = "Proceedings of the 54th Annual Meeting of the Association for Computational Linguistics (Volume 1: Long Papers)",month = aug,year = "2016",address = "Berlin, Germany",publisher = acl,url = anth # {P16-1009/},doi = "10.18653/v1/P16-1009",pages = "86--96"}

@inproceedings{kim-rush-2016-sequence,title = "Sequence-Level Knowledge Distillation",author = "Kim, Yoon and Rush, Alexander M.",editor = "Su, Jian and Duh, Kevin and Carreras, Xavier",booktitle = "Proceedings of the 2016 Conference on Empirical Methods in Natural Language Processing",month = nov,year = "2016",address = "Austin, Texas",publisher = acl,url = anth # {D16-1139/},doi = "10.18653/v1/D16-1139",pages = "1317--1327"}

@inproceedings{tiedemann-2012-parallel,title = "Parallel Data, Tools and Interfaces in {OPUS}",author = {Tiedemann, J{\"o}rg},editor = "Calzolari, Nicoletta and Choukri, Khalid and Declerck, Thierry and Do{\u{g}}an, Mehmet U{\u{g}}ur and Maegaard, Bente and Mariani, Joseph and Moreno, Asuncion and Odijk, Jan and Piperidis, Stelios",booktitle = "Proceedings of the Eighth International Conference on Language Resources and Evaluation ({LREC}'12)",month = may,year = "2012",address = "Istanbul, Turkey",publisher = "European Language Resources Association (ELRA)",url = anth # {L12-1246/},pages = "2214--2218"}

\section*{Appendix}
\appendix
\section{Derivation of the Allocation Rule}
\label{app}

To solve the optimization problem from Equation~\ref{eq}, define
\begin{align}
    &C_\mu(\lambda) \equiv \mathbb{E}[K|\mu],\\
    &G_\mu(\lambda) \equiv \mathbb{E}[X_K\mid\mu] = \frac{\int_\lambda^\infty\mathrm{d}x\, x f_\mu(x), }{ 1-F_\mu(\lambda) }.
\end{align}
Setting the constraint via a Lagrange multiplier $\Lambda$ results in the functional
\begin{align}
    \mathcal{L}[\lambda] = \mathbb{E}_{\mu}\left[G_\mu(\lambda(\mu))\right] - \Lambda \bigl(\mathbb{E}_{\mu}\left[C_\mu(\lambda(\mu))\right] - N\bigr).
\end{align}
The corresponding first-order condition is
\begin{align}
\frac{\delta\mathcal{L}}{\delta \lambda(\mu)} = G_\mu'(\lambda(\mu)) - \Lambda C_\mu'(\lambda(\mu)) = 0,
\end{align}
where
\begin{align}
    G_\mu'(\lambda) &= \frac{f_\mu(\lambda)}{1-F_\mu(\lambda)}(G_\mu(\lambda) - \lambda),\\
    C_\mu'(\lambda) &= \frac{f_\mu(\lambda)}{(1-F_\mu(\lambda))^2}.
\end{align}
This yields the equation
\begin{align}
    \Lambda &= (1-F_\mu(\lambda))(G_\mu(\lambda)-\lambda)\\
    &= \int_\lambda^\infty\mathrm{d}x (x-\lambda) f_\mu(x)\\
    &= \mathbb{E}[(X - \lambda(\mu))_+| \mu]
\end{align}

\end{document}